\documentclass[letterpaper]{article} 
\usepackage{aaai25}  
\usepackage{times}  
\usepackage{helvet}  
\usepackage{courier}  
\usepackage[hyphens]{url}  
\usepackage{graphicx} 
\usepackage{natbib}  
\usepackage{caption} 
\usepackage{algorithm}
\usepackage{algorithmic}

\usepackage{newfloat}
\usepackage{listings}
\usepackage{multirow}
\usepackage{adjustbox}
\usepackage{array}
\usepackage{amsmath}
\usepackage{color}
\usepackage{colortbl}
\usepackage{amssymb}
\usepackage{booktabs}
\usepackage{xcolor}  
\usepackage{color}
\usepackage[linesnumbered,ruled,vlined,algo2e]{algorithm2e}
\DeclareCaptionStyle{ruled}{labelfont=normalfont,labelsep=colon,strut=off} 
\floatstyle{ruled}
\newfloat{listing}{tb}{lst}{}
\floatname{listing}{Listing}
\title{ActPrompt: In-Domain Feature Adaptation via \\ Action Cues for Video Temporal Grounding}
\author {
    Yubin Wang\textsuperscript{\rm 1},
    Xinyang Jiang\textsuperscript{\rm 2},
    De Cheng\textsuperscript{\rm 3},
    Dongsheng Li\textsuperscript{\rm 2},
    Cairong Zhao\protect\textsuperscript{\rm 1}\thanks{Corresponding Author (zhaocairong@tongji.edu.cn)}
}
\affiliations {
    \textsuperscript{\rm 1}Tongji University\\
    \textsuperscript{\rm 2}Microsoft Research Asia\\
    \textsuperscript{\rm 3}Xidian University\\
}

\begin{document}

\maketitle

\begin{abstract}
Video temporal grounding is an emerging topic aiming to identify specific clips within videos.
In addition to pre-trained video models, contemporary methods utilize pre-trained vision-language models (VLM) to capture detailed characteristics of diverse scenes and objects from video frames. However, as pre-trained on images, VLM may struggle to distinguish action-sensitive patterns from static objects, making it necessary to adapt them to specific data domains for effective feature representation over temporal grounding.
We address two primary challenges to achieve this goal. 
Specifically, to mitigate high adaptation costs, we propose an efficient preliminary in-domain fine-tuning paradigm 
for feature adaptation, where downstream-adaptive features are learned through several pretext tasks. Furthermore, to integrate action-sensitive information into VLM, we introduce \textbf{A}ction-\textbf{C}ue-Injected \textbf{T}emporal \textbf{Prompt} Learning (ActPrompt), which injects action cues into the image encoder of VLM for better discovering action-sensitive patterns. Extensive experiments demonstrate that ActPrompt is an off-the-shelf training framework that can be effectively applied to various SOTA methods, resulting in notable improvements.
The complete code used in this study is provided in the supplementary materials.
\end{abstract}
\section{Introduction}
\label{sec:intro}
\begin{figure}[!t]
    \centering
    \center{\includegraphics[width=8.4cm]{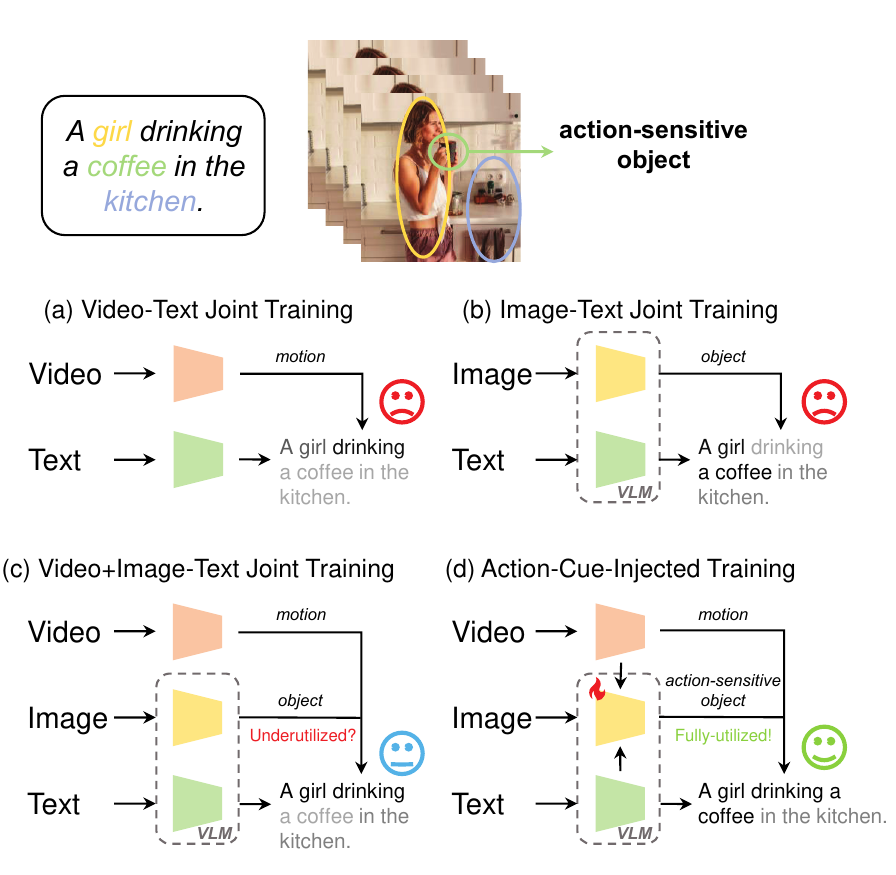}}
    \caption{Illustration of feature extraction pipelines. The image encoder in VLM can capture more detailed features of static objects, but may not be able to distinguish action-sensitive objects from backgrounds. 
    Action cues from other modalities are necessary to guide the image encoder in recognizing action-sensitive objects, making it fully utilized.\label{fig:1}}
\end{figure}

As video has emerged as a dominant medium in our daily lives, the time-consuming nature of video inspection impedes capturing desired moments or highlights~\cite{yan2023unloc, apostolidis2021video}. 
Learning high-quality feature representations for videos is essential for video temporal grounding. This task includes two crucial research areas: video moment retrieval~\cite{zhang2020learning, zhang2020span, mun2020local}, which focuses on localizing temporal windows with natural sentences, and video highlight detection~\cite{badamdorj2022contrastive, wei2022learning}, aimed at identifying key segments with highest worthiness.
Most methods~\cite{gao2021fast, gao2021relation, xiao2021boundary} employ pre-trained models as the feature extractor, exhibiting good semantic expressions within their respective modalities. 
As shown in Figure \ref{fig:1}(a), conventional methods mainly adopt pre-trained video encoders to capture motions or actions in the video. 
Vision-Language Models (VLM)~\cite{jia2021scaling, radford2021learning} are proved to be another crucial type of feature extractors for temporal grounding, due to their ability to capture detailed characteristics of diverse scenes or objects (shown in Figure \ref{fig:1}(b)). 
However, being pre-trained on massive image-text data, VLM tend to focus on the semantics of static objects. The domain gap between image datasets used for pre-training and downstream video datasets may cause VLM to overlook action-sensitive patterns (e.g., objects that are moving or being moved), crucial for temporal grounding tasks~\cite{hendricks2021probing}. 
For example, in the video segment in Figure  
\ref{fig:1}, where a lady is drinking coffee, VLM should focus on the coffee mug and the hand holding it, as both are crucial objects to recognize the action of `drinking'. 
Some works~\cite{lei2021detecting, moon2023query} attempt to compensate the motion information by fusing VLM features with video encoder features (shown in Figure \ref{fig:1}(c)). We argue that this late fusion is insufficient for deeply integrating spatial and temporal information, and the potential of adapting VLM's image encoder for recognizing action-sensitive patterns urgently needs to be explored.

However, adapting the pre-trained VLM to domain-specific video temporal grounding tasks and empowering it to recognize action-sensitive patterns is non-trivial due to two major challenges. 
Firstly, trivial end-to-end fine-tuning is impractical because it requires training the feature encoder and the downstream model concurrently on long raw videos, resulting in extremely high computational overhead.
Consequently, we introduce an efficient in-domain feature adaptation strategy. This strategy aims to fine-tune VLM's image encoder using in-domain data to produce more adaptive features as a preliminary step before standard downstream grounding.
Instead of pre-training on external large-scale datasets that may be out-of-domain, we propose a novel prompt learning algorithm that only trains a small fraction of parameters on downstream datasets.
Two pretext training tasks are jointly proposed to excavate temporal information, including moment-query pairwise ranking and moment-query contrastive learning.
After efficiently adapting VLM to downstream tasks through fine-tuning the image encoder, features are then pre-extracted as input to train downstream task-specific models, which still maintain inherent knowledge from VLM.

Secondly, existing pre-trained VLM for video temporal grounding are designed to encode individual frames and cannot model motions or actions well, making adaptation to datasets with rich temporal cues challenging.
As a result, we propose \textbf{A}ction-\textbf{C}ue-Injected \textbf{T}emporal \textbf{Prompt} Learning (ActPrompt), thereby equipping the image encoder with the capability to model temporal information (shown in Figure \ref{fig:1}(d)). 
ActPrompt contains two main modules. First, Action Cue Injection module (ACI) facilitates VLM to capture action-sensitive patterns from images under the guidance of action cues from other modalities.
Features from the video encoder and the text encoder are utilized for producing action cues, which are injected into the image encoder in the form of prompt embeddings, assisting VLM in focusing on action-sensitive regions in static images.
After identifying action-sensitive visual regions via attention scores in individual images, we introduce Context-aware Temporal Prompt Learning (CTPL), which extracts motion features from a temporal sequence of selected visual regions in consecutive frames, considering the temporal context.
An additional adaptor is trained to acquire temporal prompts from motion features, incorporating temporal knowledge into a static video frame with minimal computational overhead.

The contributions of our work are summarized as follows.
1) We introduce an efficient in-domain fine-tuning training strategy for VLM's image encoder before standard downstream grounding. Two pretext training tasks are jointly proposed to excavate temporal information from in-domain data.
2) We propose Action-Cue-Injected Temporal Prompt Learning (ActPrompt) as an off-the-shelf framework to facilitate VLM modeling actions and motions. ACI assists VLM in capturing action-related patterns from images via action cues, while CTPL extracts motion features from visual regions in the temporal context.
3) Our experiments on both moment retrieval and highlight detection improve all baselines and validate the significance of downstream-adaptive features for video temporal grounding.

\section{Related works}
\begin{figure*}[!t]
    \centering    \center{\includegraphics[width=17.5cm]{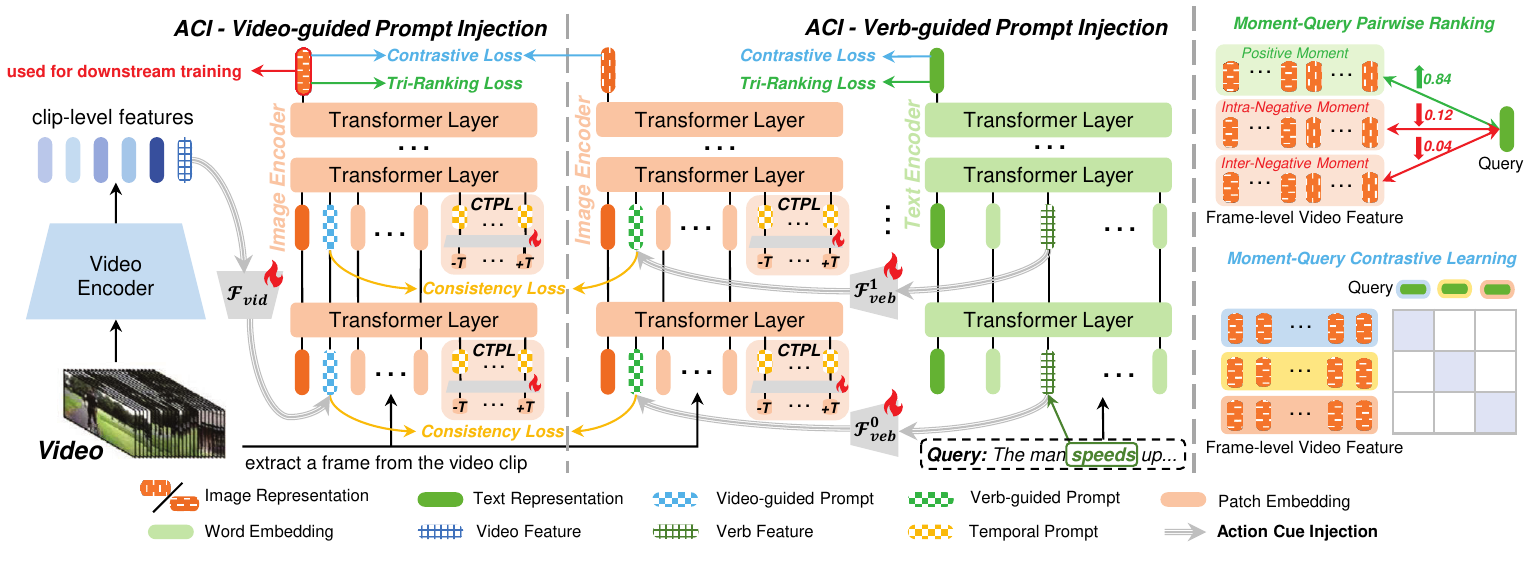}}
    \caption{The overall framework of ActPrompt. To capture visual regions related to motions, Action Cue Injection (ACI) injects video- and verb-guided prompts from other encoders into VLM's image encoder as action cues.  
    Context-aware Temporal Prompt Learning (CTPL) selects action-sensitive visual regions from consecutive frames via ACI and groups them to generate temporal prompt. 
    The output representations are fed into training objectives of pretext tasks including moment-query pairwise ranking and moment-query contrastive learning for adaption to downstream grounding. The fine-tuned modules are marked with a red flame pattern, with the other modules frozen.\label{fig:3}}
\end{figure*} 
\subsection{Video temporal grounding}
We examine two primary tasks in video temporal grounding: moment retrieval and highlight detection, comparing them as distinct variations of a shared problem. For moment retrieval, existing studies assume single~\cite{anne2017localizing, gao2017tall} or multiple~\cite{lei2020tvr} continuous moments in a video corresponding to a given text query, typically focused on activities. Furthermore, some studies~\cite{zhang2020learning, zhang2019man, xu2019multilevel} attempt to score generated moment proposals.
In contrast, alternative methods~\cite{mun2020local, li2021proposal, xu2022hisa} directly regress the start and end boundaries without the need for proposal candidates.
For highlight detection, the existing datasets~\cite{song2015tvsum, sun2014ranking} are typically domain-specific and query-agnostic, leading to various approaches~\cite{hong2020mini, xu2021cross, badamdorj2021joint} that treat the task as a scoring problem.
To tackle both tasks concurrently, Moment-DETR~\cite{lei2021detecting} introduces QVHighlights dataset and also presents a modified version of the detection Transformer to pinpoint query-relevant moments and their associated saliency scores. 
Subsequently, other works~\cite{liu2022umt, moon2023query, lin2023univtg, sun2024tr} concentrate on processing multi-modal data based on Transformer-based architectures. 
Based on these methods, we provide an in-domain fine-tuning paradigm for VLM to learn adaptive features by injecting action cues into the image encoder.

\subsection{Prompt learning}
Recent prompt learning methods~\cite{li2021prefix, lester2021power, liu2021p} utilize continuous contexts to automate prompt engineering and explore optimal prompts. This paradigm can also be extended to vision-language models~\cite{radford2021learning, jia2021scaling} by recent works~\cite{zhou2022learning,zhou2022conditional,lu2022prompt,zhu2023prompt}. 
Notably, as a pioneer work, CoOp~\cite{zhou2022learning} demonstrates that a suitable prompt for enhancing the recognition performance of CLIP can be learned with very few samples. 
To explore the potential of learning prompts for visual modality, Visual Prompt Tuning (VPT)~\cite{jia2022visual} introduces only a minimal set of trainable vectors as prompts, while keeping the model backbone frozen. 
Other multi-modal prompt learning approaches~\cite {khattak2022maple,zhao2024learning,wang2023learning} also attempt to enhance the alignment between representations from vision and text modalities. 
Instead of focusing on classification, we introduce two kinds of novel prompts including action-aware prompt and temporal prompt for video temporal grounding, aiming to facilitate VLM modeling motions with prior temporal knowledge.

\section{Method}
\subsection{Overview}
In this section, we first provide an overview of the general feature extraction paradigm for video temporal grounding, which leverages CLIP~\cite{radford2021learning} as the image encoder and text encoder, and SlowFast~\cite{feichtenhofer2019slowfast} as the video encoder.
Given a video consisting of $L$ clips $\{c_1, c_2, ..., c_L\}$ and a text query with $N$ words $\{w_1, w_2, ..., w_N\}$, we extract video feature $v^V_{t}\in \mathbb{R}^{D_V}$ and image feature $v^I_{t}\in \mathbb{R}^{D_I}$ of the clip $c_t$ at the $t$-th timestamp using the video encoder and the image encoder respectively. 
Notably, the 2D frame $c_t^{'}$ extracted from $c_t$ is used as input of the image encoder. 
Additionally, we extract textual feature $t_i\in \mathbb{R}^{D_I}$ of the $i$-th word token in the text query via the text encoder, which has the same feature dimension as $v^I_{t}$.
The image feature and the video feature are fused as the video representation $v_t \in \mathbb{R}^{D_I}$, and then clip-level video representations $V = \{v_1, v_2, ..., v_L\}$ and token-level text representations $T = \{t_1, t_2, ..., t_N\}$ can be used for downstream tasks.
Instead of regularly fine-tuning all encoders in the above feature extraction pipeline, we propose Action-Cue-Injected Temporal Prompt Learning (ActPrompt), specifically for fine-tuning the image encoder in VLM with action cues from downstream datasets.
The overall framework is illustrated in Figure \ref{fig:3}.

\subsection{Action cue injection}
\label{subsec:action}
Action cues refer to the signal that indicates an object to take specific actions or engage in particular behaviors. 
We develop Action Cue Injection (ACI) that encapsulates action cues from other modalities as action-aware prompt, which are then incorporated into the image encoder to guide it to focus on visual regions related to motions and actions. We mainly introduce two types of prompts: video-guided prompt and verb-guided prompt. We feed the two prompts into the image encoder separately to produce two representations of one frame in parallel.

\paragraph{Video-guided prompt} Given a specific clip $c_t$, we feed its video feature $v^V_{t}$ from the video encoder into a video-image coupling function $\mathcal{F}_{vid}(\cdot)$, implemented as a linear layer, to produce video-guided prompt $p_{vid,t}=\mathcal{F}_{vid}(v^V_{t}) \in \mathbb{R}^D$, where $D$ denotes the dimension of patch embeddings
This prompt is then concatenated with patch embeddings at the first layer of the image encoder. 
Through self-attention in each Transformer layer, this prompt interacts with patch embeddings and learns to pay more attention to regions with action patterns through our training objectives. 

\paragraph{Verb-guided prompt} We leverage a language parsing tool\footnote{spaCy: https://spacy.io/} to extract the verb token $w_{v} \in \{w_1, w_2, ..., w_N\}$ from the text query, where $v$ denotes its index in the word sequence. 
Next, we obtain the embedding associated with that verb token from each layer of the text encoder as verb features.
As it passes through Transformer layers, the embedding interacts with other word embeddings via self-attention, thus complementing semantic contextual information, such as action-sensitive objects beyond the verb itself.
Given the embedding $e^{l}_{v}$ of the $l$-th layer's output related to $w_{v}$, we feed it into a layer-specific verb-image coupling function $\mathcal{F}^{l}_{veb}(\cdot)$, which is also implemented as a linear layer, to produce verb-guided prompt $p^{l}_{veb}=\mathcal{F}^{l}_{veb}(e^{l}_{v}) \in \mathbb{R}^D$. 
This prompt is then concatenated with patch embeddings at the $l$-th layer of the image encoder.
The absence of subscript $t$ here is because it is not linked to any specific video frame clip $c_t$. 
Due to the structural similarity between the image encoder and the text encoder, the image encoder can receive supervisory textual signals at each layer.

\paragraph{Consistency learning} 
Despite being generated from different modalities, they are expected to exhibit consistent semantics towards actions, where the patch-wise attention distribution from them should be close.
In this regard, we introduce a consistency loss to reduce the disparity between patch-wise attention scores derived from the two prompts, aiming to attain unified concepts without ambiguity.
Given a video containing $L$ clips, we denote the patch-wise attention map (mean by attention heads) of prompts $p^l_{vid, t}$ and $p^l_{veb}$ at the $l$-th layer for the $t$-th frame as $\mathbf{A}^l_{vid, t} \in \mathbb{R}^{N_p}$ and $\mathbf{A}^l_{veb} \in \mathbb{R}^{N_p}$, where $N_p$ denotes the number of patches. We use Mean Square Error (MSE) to measure the difference as follows:
\begin{equation}
\label{eq:1}
\mathcal{L}_{con}=\frac{1}{L}\sum_{t=1}^{L}\sum_{l=1}^{N_L}\operatorname{MSELoss}(\mathbf{A}^l_{vid}, \mathbf{A}^l_{veb, t}), 
\end{equation}
where $N_L$ denotes the number of Transformer layers. 

\begin{figure}[!t]
    \centering
\center{\includegraphics[width=8.5cm]{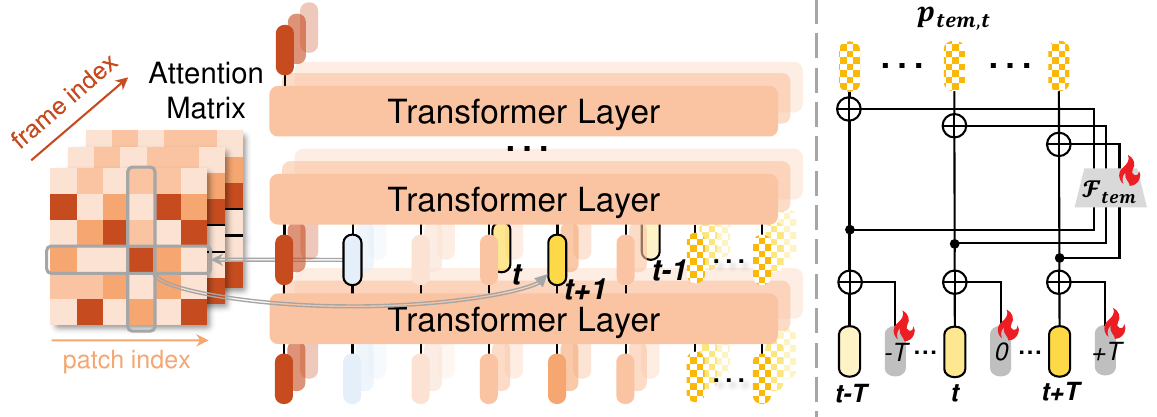}}
    \caption{Illustration of context-aware temporal prompt learning. We sample the patch with the highest attention score to the action-sensitive prompt for each frame (left) and concatenate the sampled patch embeddings from the current frame and neighboring frames for learning context-aware temporal prompt (right).\label{fig:4}}
\end{figure} 
\subsection{Context-aware temporal prompt learning}
\label{subsec:context}
VLM also suffer from a deficiency in modeling temporal information from consecutive frames.
Consequently, we propose Context-aware Temporal Prompt Learning (CTPL) to enable VLM to extract temporal information from the current frame and its neighboring frames, as shown in Figure \ref{fig:4}. We take video-guided prompt as an example, where the process of applying verb-guided prompt is similar.

Given the concept-consistent attention map $\mathbf{A}^l_{vid, t}$ via ACI, we sample the patch $\tilde{e}^l_{t}$ with the highest attention score to the corresponding prompt at the $l$-th layer in the $t$-th frame as $\tilde{e}^l_{t} = E_{l,t}[i^*] \in \mathbb{R}^{D}$, where$\ i^*=\arg\max_{i} \mathbf{A}^l_{vid, t}[i]$, and $E_{l,t}$ indicates the patch embeddings of the $t$-th frame at the $l$-th layer.
After sampling the patch from the current image, we concatenate $\tilde{e}^l_{t}$ with other selected embeddings from its previous and subsequent $T$ frames at the first dimension to form a temporal sequence of patches. Here we apply a replication padding for boundary frames when conducting CTPL. 
We add layer-specific positional embeddings $P^l \in \mathbb{R}^{(2T+1) \times D}$ to obtain embeddings $\tilde{E}^l_{tem, t} =[\tilde{e}^l_{t-T}, ..., \tilde{e}^l_{t}, ..., \tilde{e}^l_{t+T}] + P^l$, where $[\cdot,\cdot]$ stands for concatenation at the first dimension. $\tilde{E}^l_{tem, t}$ are then fed into an MLP-based temporal prompt generator $\mathcal{F}_{tem}(\cdot)$ with a residual shortcut to produce context-aware temporal prompt.
Hence, for the $t$-th frame at the $l$-th layer, CTPL obtain the temporal prompt $p^l_{tem, t} \in \mathbb{R}^{(2T+1) \times D}$ as $p^l_{tem, t}  = \mathcal{F}_{tem}(\tilde{E}^l_{tem, t}) + \tilde{E}^l_{tem, t}$, which is then inserted into the ($l$+1)-th Transformer layer with other embeddings. The process is formulated as:

\begin{align} {\left[x_1, p_{vid}^{1}, E_1\right] } & =\operatorname{Layer}_1\left(\left[{x_0}, p_{vid}, E_0\right]\right), \nonumber\\{\left[x_{l+1}, p_{vid}^{l+1}, E_{l+1}, \_\right] } & =\operatorname{Layer}_{l+1}\left(\left[x_{l}, p_{vid}^{l}, E_{l}, p_{tem}^l\right]\right),\nonumber\\
l & = 1, 2, ..., N-1,  \label{eq:4}\end{align} 

where $x_l$ denotes the class embedding at the $l$-th layer. Here we emit the subscript $t$ since we conduct the same process for all frames. 
Notably, when using $p_{veb}^{l}$ instead of $p_{vid}^{l}$ in Equation \ref{eq:4} during verb-guided prompt injection, we replace the output embedding of the ($l$+1)-th layer at the same position with $p_{veb}^{l}$, using the new learned prompt embedding from the output of $\mathcal{F}^{l}_{veb}(\cdot)$ as the input for the next layer.
We consider Equation \ref{eq:4} as the final form for our approach, wherein ACI and CTPL operate concurrently and mutually enhance each other. ACI supplies action-aware prompts $p_{vid}$ and $p_{veb}$, emphasizing action-sensitive regions with patch-wise attention. Meanwhile, CTPL utilizes auxiliary cues from neighboring frames to generate the temporal prompt $p_{tem}$, enhancing temporal modeling by considering the temporal context.
We then have frame-level video representations $V^I_{vid}=\{v^I_{vid,t}\}_{t=1}^L$ based on video-guided prompt and $V^I_{veb}=\{v^I_{veb,t}\}_{t=1}^L$ based on verb-guided prompt.
\begin{table*}[t]
\centering
\footnotesize
\setlength{\tabcolsep}{9pt}
\begin{tabular}{lccccccc}
    \toprule
    \multirow{3}*{\textbf{Method}} & \multicolumn{5}{c}{\textbf{Moment Retrieval}} & \multicolumn{2}{c}{\textbf{HD}} \\
    \cmidrule(lr){2-6}\cmidrule(lr){7-8}
    ~ & \multicolumn{2}{c}{R1} & \multicolumn{3}{c}{mAP} & \multicolumn{2}{c}{$\geq$ Very Good} \\ 
    \cmidrule(lr){2-3}\cmidrule(lr){4-6}\cmidrule(lr){7-8}
    ~ & @0.5 & @0.7 & @0.5 & @0.75 & Avg. & mAP & HIT@1 \\
    \midrule
    CAL~\cite{escorcia2019temporal} & 25.49 & 11.54 & 23.40 & 7.65 & 9.89 & - & - \\ %
    XML~\cite{lei2020tvr} & 41.83 & 30.35 & 44.63 & 31.73 & 32.14 & 34.49 & 55.25 \\ %
    XML+~\cite{lei2020tvr} & 46.69 & 33.46 & 47.89 & 34.67 & 34.90 & 35.38 & 55.06 \\ %
    UMT~\cite{liu2022umt} &  56.23 & 41.18 & 53.83 & 37.01 & 36.12 & 38.18 & 59.99 \\ %
    M-DETR~\cite{lei2021detecting} & 52.89 & 33.02 & 54.82 & 29.40 & 30.73 & 35.69 & 55.60 \\ %
    UniVTG~\cite{lin2023univtg} & 58.86 & 40.86 & 57.60 & 35.59 & 35.47 & 38.20 & 60.96\\ %
    QD-DETR~\cite{moon2023query} & 62.40 & 44.98 & 62.52 & 39.88 & 39.86 & 38.94 & 62.40 \\ %
    TR-DETR~\cite{sun2024tr} & 64.66 & 48.96 & 63.98 & 43.73 & 42.62 & 39.91 & 63.42 \\
    \midrule
    ActPrompt (w/ M-DETR) & 59.34 & 39.75 & 59.87 & 34.76 & 35.71 & 37.29 & 59.66 \\
    ActPrompt (w/ UniVTG) & 60.57 & 42.22 & 59.72 & 37.86 & 37.65 & 38.82 & 62.71 \\
    ActPrompt (w/ QD-DETR) & 62.58 & 46.89 & 62.41 & 41.42 & 41.29 & 40.02 & 64.46 \\
    ActPrompt (w/ TR-DETR) & \textbf{65.28} & \textbf{49.76} & \textbf{64.91} & \textbf{44.02} & \textbf{43.39} & \textbf{40.72} & \textbf{65.01} \\
    \bottomrule
\end{tabular}
\caption{Jointly moment retrieval and highlight detection results on QVHighlights test split.\label{tab:qv}}
\end{table*}
\subsection{Pretext tasks for adaption to downstream grounding}
\label{subsec:pretext}
To facilitate adapting VLM to video temporal grounding, we design two pretext tasks as training objectives, including moment-query pairwise ranking and moment-query contrastive learning. These tasks ensure an improved acquisition of action-sensitive semantics in aforementioned modules.

Given a moment-query pair $(m, q)$ randomly selected from a training data batch, to facilitate moment-query pairwise ranking, we generate a non-overlap moment from the same video of $m$ as an intra-video negative sample $m^{intra-}$. 
Furthermore, we identify another moment, $m^{inter-}$, from other videos in the batch to act as an inter-video negative moment. This process results in a quadruple $(m^{+}, m^{intra-}, m^{inter-}, q)$, where we rename the original moment $m$ as $m^{+}$ to maintain uniformity in expression format.
Each moment in the quadruple is represented by two sets of frame-level features $V^I_{vid}$ and $V^I_{veb}$, obtained by injecting video-guided or verb-guided prompt in ACI separately. 
We normalize and average two sets of frame-level representations of these moments respectively to obtain moment-level representations $\tilde{V}_{vid}=\{\tilde{v}^{+}_{vid},\tilde{v}^{intra-}_{vid},\tilde{v}^{inter-}_{vid}\}$ and $\tilde{V}_{veb}=\{\tilde{v}^{+}_{veb},\tilde{v}^{intra-}_{veb},\tilde{v}^{inter-}_{veb}\}$.
Notably, verb-guided prompt $p_{veb}$ for all moments within the quadruple is generated via verb features from the text query $q$.
Given the representation of the text query $q$, denoted as $t^q$ as an anchor, our goal is to have it aligned closely to the positive moment and far from the negative ones.
As a result, we propose triplet-ranking loss for moment-query pairwise ranking, formulated as:
\begin{align}
\mathcal{L}_{tri} = & -\log (\frac{\operatorname{exp}(\operatorname{sim}(\tilde{v}^{+}_{vid},{t^q}))}{\sum_{\tilde{v} \in \tilde{V}_{vid}}\operatorname{exp}(\operatorname{sim}(\tilde{v},{t^q})))} \notag \\
& \cdot \frac{\operatorname{exp}(\operatorname{sim}(\tilde{v}^{+}_{veb},{t^q}))}{\sum_{\tilde{v} \in \tilde{V}_{veb}}\operatorname{exp}(\operatorname{sim}(\tilde{v},{t^q})))}),
\end{align}
where $\operatorname{sim}(\cdot,\cdot)$ is the cosine similarity metrics.

Moreover, we conduct moment-query contrastive learning to align positive pairs using cross-entropy loss. Since representations associated with verb-guided prompt have already contained information from the corresponding query and thus are not suitable for this process, we only leverage $\tilde{v}^{+}_{vid}$ as the moment-level representation for the positive moment. The cross-entropy loss is formulated as follows:
\begin{align}
    \mathcal{L}_{ce} = -\log\frac{\operatorname{exp}({\operatorname{sim}(\tilde{v}^{+}_{vid}, t^q))}}{\sum_{\tilde{t} \in T_q} \operatorname{exp}({\operatorname{sim}(\tilde{v}^{+}_{vid}, \tilde{t})})}.
\end{align}
Here $T_q$ denotes the set of query representations within the batch where $t^q \in T_q$.

The total loss function can be expressed as follows:
\begin{equation}
    \mathcal{L}_{total} = \mathcal{L}_{ce}+\alpha_{1}\mathcal{L}_{tri}+\alpha_{2}\mathcal{L}_{con},
\end{equation}
where $\alpha_{1}$ and $\alpha_{2}$ are coefficients for balancing these losses.
We finally leverage $V^I_{vid}$ as image features for downstream tasks, since text queries are clip-agnostic and not accessible during clip-level video feature extraction.

\section{Experiments}
\subsection{Datasets and settings}
\paragraph{Datasets} 
QVHighlights~\cite{lei2021detecting} is a public dataset with ground-truth annotations for both moment retrieval and highlight detection. It contains 10,148 short video segments, each annotated with at least one text query indicating its relevant moments. Evaluation on the test split can only be conducted by submitting predictions to the QVHighlights server, ensuring a fair benchmark. Charades-STA~\cite{gao2017tall} contains 16,128 indoor videos with an average duration of 30.6 seconds, divided into 12,408 query-interval pairs for training and 3,720 pairs for testing. TACoS~\cite{regneri2013grounding} consists of 127 videos with an average duration of 4.78 minutes, where 75, 27 and 25 videos are for training, validation, and testing.
\paragraph{Evaluation metrics}
For QVHighlights, we follow official metrics in~\cite{lei2021detecting}, where Recall@1 with IoU thresholds 0.5 and 0.7, mAP with IoU thresholds 0.5 and 0.75, and the average mAP over a series of IoU thresholds are used for moment retrieval. For highlight detection, mAP and HIT@1 are used, where a true positive clip has a saliency score of Very Good. For Charades-STA and TACoS, Recall@1 with IoU thresholds 0.3, 0.5 and 0.7, and mIoU are used following works~\cite{lei2021detecting, moon2023query, lin2023univtg, sun2024tr}.
\begin{table*}[!t]
\centering
\small
\setlength{\tabcolsep}{6.3pt}
\begin{tabular}{l|cccc|cccc}
    \toprule
    \multirow{2}*{\textbf{Method}} & \multicolumn{4}{c|}{\textbf{Charades-STA}} & \multicolumn{4}{c}{\textbf{TACoS}}\\
    ~ & R@0.3 & R@0.5 & R@0.7 & mIoU & R@0.3 & R@0.5 & R@0.7 & mIoU \\ 
    \midrule
    2D-TAN~\cite{zhang2020learning} & 58.76 & 46.02 & 27.50 & 41.25 & 40.01 & 27.99 & 12.92 & 27.22 \\ %
    VSLNet~\cite{zhang2020span} & 60.30 & 42.69 & 24.14 & 41.58 & 35.54 & 23.54 & 13.15 & 24.99  \\ %
    M-DETR~\cite{lei2021detecting} & 65.83 & 52.07 & 30.59 & 45.54 & 37.97 & 24.67 & 11.97 & 25.49  \\ %
    QD-DETR~\cite{moon2023query} & 69.81 & 57.39 & 34.78 & 49.41 & 42.01 & 30.56 & 17.09 & 28.27  \\ %
    TR-DETR~\cite{sun2024tr} & 69.58 & 57.61 & 33.52 & 49.87 & - & - & - & - \\
    UniVTG~\cite{lin2023univtg} & 70.81 & 58.01 & 35.65 & 50.10 & 51.44 & 34.97 & 17.35 & 33.60  \\ %
    \midrule
    ActPrompt (w/ M-DETR) & 69.27 & 56.51 & 33.25 & 48.73 & 39.89 & 26.69 & 13.65 & 26.04  \\
    ActPrompt (w/ QD-DETR)  & \textbf{72.90} & 59.95 & 37.58 & \textbf{51.66} & 43.41 & 32.77 & \textbf{19.17} & 29.43  \\
    ActPrompt (w/ TR-DETR)  & 72.21 & 59.84 & 37.32 & 51.23 & - & - & - & - \\
    ActPrompt (w/ UniVTG)  & 71.59 & \textbf{60.35} & \textbf{38.12} & 51.31 & \textbf{52.64} & \textbf{36.77} & 17.12 & \textbf{34.24}  \\
    \bottomrule
\end{tabular}
\caption{Moment retrieval results on Charades-STA, and TACoS benchmarks.\label{tab:mom}}
\end{table*}
\paragraph{Implementation details}
 Following~\cite{lei2021detecting, moon2023query, lin2023univtg, sun2024tr}, we leverage CLIP~\cite{radford2021learning} (ViT-B/32) as the image and text encoder, and SlowFast~\cite{feichtenhofer2019slowfast} (ResNet-50) as the video encoder. 
 We set an initial learning rate of 0.1 and train our model for 10 epochs. The neighboring frame number is set to $T=1$. The coefficients of loss function are set to $\alpha_{1}=5$ (for TACoS and QVHighlights) or $\alpha_{1}=10$ (for Charades-STA) and $\alpha_{2}=200$. We use a data ratio of 0.1 for each epoch in a disjoint manner, thus forcing each in-domain sample to be utilized at our fine-tuning stage. The network is trained using a single NVIDIA RTX 3090 GPU, requiring no more than a few hours per specific dataset. 
The total number of trainable parameters is 12M, which is relatively small compared to the 86M parameters required to train the entire image encoder.
 We introduce several SOTA methods in downstream tasks as baselines, and conduct our ActPrompt on some of them, including Moment-DETR~\cite{lei2021detecting}, QD-DETR~\cite{moon2023query}, 
 UniVTG~\cite{lin2023univtg} and TR-DETR~\cite{sun2024tr}. All baselines use the same video and text features on a specific dataset.
\subsection{Joint moment retrieval and highlight detection}
As illustrated in Table \ref{tab:qv}, we evaluate our approach on the QVHighlights test split, alongside baselines that use features from a frozen extractor. 
Our method has shown remarkable performance gains and demonstrated superiority in in-domain fine-tuning. 
Specifically, results of ActPrompt with Moment-DETR exhibit a significant improvement, such as +4.98\% Avg. mAP in moment retrieval and +4.06\% HIT@1 in highlight detection over frozen features. 
This validates that our method guarantees a strong performance bound for video temporal grounding. 
Note that the performance gain of ActPrompt for QD-DETR is minimal at R1@0.5 and even slightly decreases at mAP@0.5. This is because the grounding performance at an IoU threshold of 0.5 relies mainly on the ability to recognize static patterns for coarse predictions, an area where QD-DETR already excels.
However, our fine-tuning strategy shows an improvement of +1.91\% at R1@0.7 and +1.54\% at mAP@0.75, which suggests that action cues assist in refining predicted bounding boxes for more accurate grounding. 
For highlight detection, the consistent improvement over baselines indicates our enhancement in capturing highlight segments. In general, our fine-tuning strategy significantly improves the state-of-the-art method, TR-DETR, to a notable extent.
\subsection{Moment retrieval}
In Table \ref{tab:mom}, we evaluate our method on two widely used moment retrieval benchmarks, TACoS and Charades-STA. Similar to the observation made by QVHighlights, our approach is still superior to all baselines. This demonstrates once more the effectiveness of our in-domain fine-tuning strategy. Our high-quality features have resulted in significant improvements, leading to a considerable increase in the mIoU i.e., +3.19\% in Charades-STA for Moment-DETR and +1.16\% in TACoS for QD-DETR. 
However, it is worth mentioning that the overall improvement over Charades-STA is greater than that of TACoS. Instead of videos with fixed scenes in TACoS, Charades-STA has videos with more diverse content and potential background disturbance, making it hard to recognize action-sensitive semantics. The deficiency in capturing non-static patterns has been effectively addressed with the guidance of action cues.

\subsection{Ablation studies}
\paragraph{Analysis on model components.} In Table \ref{tab:ab}, we investigate the influence of model components of our ActPrompt mainly from two aspects: the type of objectives and prompts. 
From an objective perspective, all loss functions positively impact the results, especially when complemented by the action-aware prompt. Regarding different prompts, it is evident that our action-aware prompt, learned through ACI, is more effective in temporal localization than the vanilla prompt, underscoring the importance of incorporating action cues into images. Moreover, the temporal prompt, learned from the temporal context through CTPL, significantly contributes to the improvement. However, the fixed scenes in TACoS somewhat hinder the recognition of useful motion information, leading to relatively smaller improvements.
\begin{figure*}[t]
    \centering
\center{\includegraphics[width=14.3cm, height=0.25\textheight]{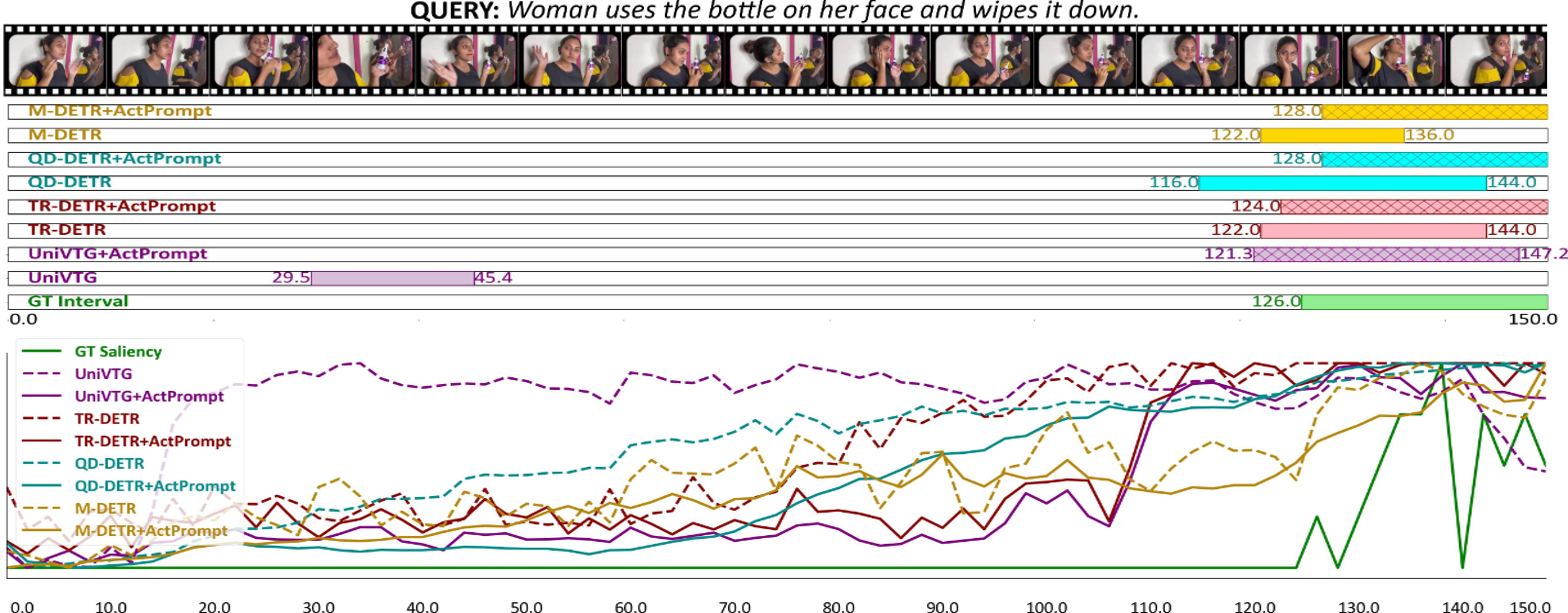}}
    \caption{Visualization of joint moment retrieval and highlight detection on QVHighlights over various baselines and their variants with our ActPrompt.\label{fig:6}}
\end{figure*} 

\begin{table}[!t]
\centering
\small
\setlength{\tabcolsep}{6.3pt}
\begin{adjustbox}{valign=t, width=1\linewidth}
\begin{tabular}{>{\centering\arraybackslash}m{0.5cm}>{\centering\arraybackslash}m{0.5cm}>{\centering\arraybackslash}m{0.5cm}|>{\centering\arraybackslash}m{0.5cm}>{\centering\arraybackslash}m{0.5cm}>{\centering\arraybackslash}m{0.5cm}|>{\centering\arraybackslash}m{0.8cm}|>{\centering\arraybackslash}m{0.8cm}|>{\centering\arraybackslash}m{0.8cm}}
        \toprule
        \multicolumn{3}{c|}{\textbf{Objective}} & \multicolumn{3}{c|}{\textbf{Prompt}} & \textbf{QVH.} & \textbf{Cha.} & \textbf{TAC.}\\
        $\mathcal{L}_{ce}$ & $\mathcal{L}_{tri}$ & $\mathcal{L}_{con}$ & Vani. & Act. & Tem. & HIT@1 & mIoU & mIoU\\ 
        \midrule
        - & - & - & - & - & - & 61.98 & 49.41 & 28.27\\
        \midrule
        $\checkmark$ & & & $\checkmark$ & & & 62.01 & 49.63 & 28.36\\
        $\checkmark$ &  &  & & $\checkmark$ &  & 62.64 & 49.94 & 28.64 \\
        $\checkmark$ & $\checkmark$ &  & & $\checkmark$ &  & 62.87 & 50.67 & 29.13 \\
        $\checkmark$ & $\checkmark$ & $\checkmark$ & & $\checkmark$ &  & 63.20 & 50.93 & 29.25\\
         $\checkmark$ & $\checkmark$ & $\checkmark$ & & $\checkmark$ & $\checkmark$ & \textbf{64.02} & \textbf{51.66} & \textbf{29.43}\\
        \bottomrule
        \end{tabular}
        \end{adjustbox}
\caption{\textbf{Analysis on model components.} We choose QD-DETR as baseline and conduct experiments with different combinations.
For QVHighlights, we report results on \textit{validation} split due to the server's submission limitation (the same for following ablations). ``Vani'' indicates vanilla prompt learning introduced in VPT~\cite{jia2022visual}.\label{tab:ab}}
\end{table}

\begin{table}[!t]
\centering
\small
\setlength{\tabcolsep}{6.3pt}
\begin{adjustbox}{valign=t, width=0.9\linewidth}
        \begin{tabular}{l|c|c|c}
            \toprule
                \textbf{Fine-Tuning} & \textbf{QVH.} & \textbf{Cha.} & \textbf{TAC.} \\
            \textbf{Method} & HIT@1 & mIoU & mIoU\\ 
            \midrule
            Frozen & 61.98 & 49.41 & 28.27 \\
            \midrule
            Video & 62.73 & 49.63 & 28.01 \\
            Adapter~\cite{houlsby2019parameter} & 63.23 & 50.38 & 28.91\\
            VPT~\cite{jia2022visual} & 63.04 & 50.50 & 28.66\\
            ActPrompt (Ours) & \textbf{64.02} & \textbf{51.66} & \textbf{29.43} \\
            \bottomrule
        \end{tabular}
        \end{adjustbox}
\caption{Analysis on fine-tuning methods.\label{tab:vid}}
\end{table}

\begin{figure}[t]
    \centering
\center{\includegraphics[width=8.3cm]{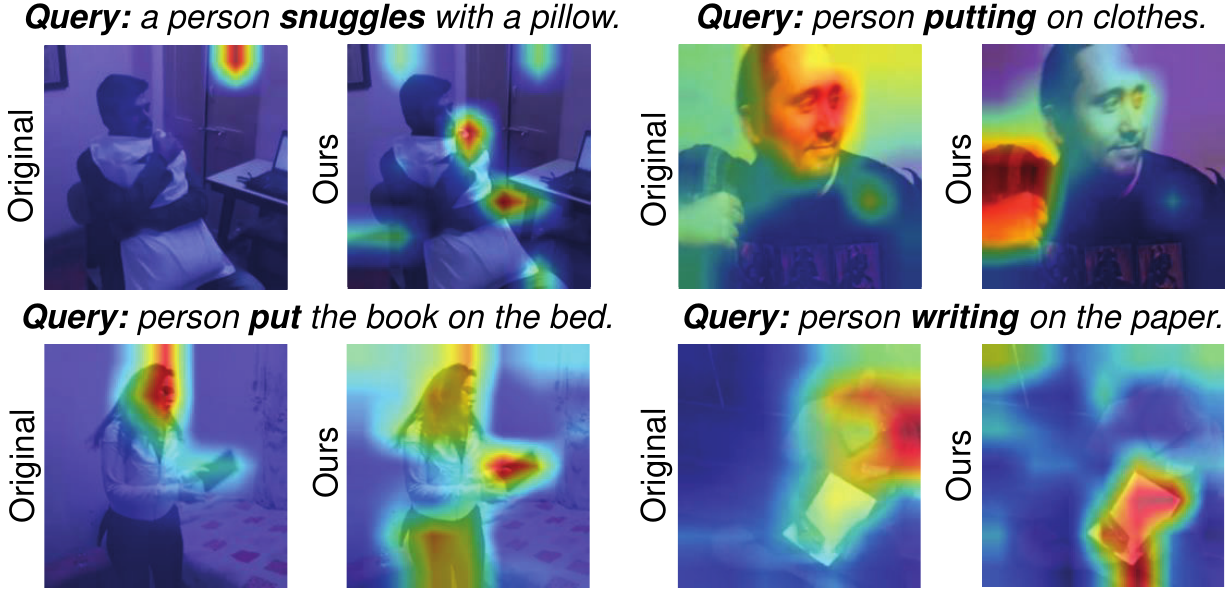}}
    \caption{Visualization of attention maps from frozen CLIP's image encoder and in-domain fine-tuned encoder.\label{fig:5}}
\end{figure} 

\paragraph{Analysis on other fine-tuning methods.} 
We compare our in-domain fine-tuning method against other fine-tuning approaches on Transformer-based image encoders. The Adapter~\cite{houlsby2019parameter} method introduces a bottleneck with relatively few parameters compared to the original model. VPT~\cite{jia2022visual} adds a minimal set of trainable vectors as prompts at each layer. We also explore the possibility of fine-tuning the video encoder by appending a learnable adapter to adjust video features. All experiments with these methods are conducted using our proposed pretext tasks to ensure a fair comparison. As shown in Table \ref{tab:vid}, the results highlight the effectiveness of action cues in fine-tuning the image encoder, significantly outperforming other methods. In contrast, fine-tuning the video encoder does not yield significant improvements and may even degrade performance when done concurrently with the image encoder. This degradation likely occurs because the video encoder is pre-trained on extensive video data, which is sufficient for generalizing to downstream tasks. Thus, further fine-tuning on the downstream dataset can result in overfitting, negatively impacting performance during testing.

\paragraph{Visualization} We provide quantitative visualizations on QVHighlights for various baselines and their variants incorporating our ActPrompt in Figure \ref{fig:6}. In moment retrieval, we observe that ActPrompt enhances the baselines by providing more accurate moment predictions, even correcting completely incorrect predictions made by UniVTG. In highlight detection, ActPrompt consistently predicts lower saliency scores for non-relevant clips compared to the baselines. We attribute this to ActPrompt's sensitivity to changes in actions or motions, resulting in improved localization performance. To further validate ActPrompt's ability to capture action-sensitive patterns, we compare the visual disparities between our image encoder and the frozen image encoder by visualizing attention maps. We randomly select several frames from various videos in Charades-STA. As shown in Figure \ref{fig:5}, our method, which incorporates action cues, alleviates the issue of focusing on action-irrelevant objects, such as faces and backgrounds, commonly observed in the frozen encoder.

\section{Conclusion}
We propose an efficient in-domain fine-tuning training strategy to adapt VLM for downstream video temporal grounding tasks with the assistance of action cues. 
Our proposed method, Action-Cue-Injected Temporal Prompt Learning (ActPrompt), enhances VLM's ability to capture action-related patterns, generating high-quality features.
However, there are still some limitations to consider. 
The efficiency and effectiveness of the in-domain fine-tuning strategy may vary across different datasets and task scenarios. 
Additionally, the performance of ActPrompt in extracting temporal information is contingent upon the quality and relevance of the action cues, which may not always be consistent or readily available in all in-domain data.
We hope more attention will be paid to the importance of action cues and better utilize them to improve VLM's recognition performance on video comprehension tasks.

\bibliography{aaai25}

\begin{thebibliography}{45}
\providecommand{\natexlab}[1]{#1}

\bibitem[{Anne~Hendricks et~al.(2017)Anne~Hendricks, Wang, Shechtman, Sivic, Darrell, and Russell}]{anne2017localizing}
Anne~Hendricks, L.; Wang, O.; Shechtman, E.; Sivic, J.; Darrell, T.; and Russell, B. 2017.
\newblock Localizing moments in video with natural language.
\newblock In \emph{Proceedings of the IEEE international conference on computer vision}, 5803--5812.

\bibitem[{Apostolidis et~al.(2021)Apostolidis, Adamantidou, Metsai, Mezaris, and Patras}]{apostolidis2021video}
Apostolidis, E.; Adamantidou, E.; Metsai, A.~I.; Mezaris, V.; and Patras, I. 2021.
\newblock Video summarization using deep neural networks: A survey.
\newblock \emph{Proceedings of the IEEE}, 109(11): 1838--1863.

\bibitem[{Badamdorj et~al.(2021)Badamdorj, Rochan, Wang, and Cheng}]{badamdorj2021joint}
Badamdorj, T.; Rochan, M.; Wang, Y.; and Cheng, L. 2021.
\newblock Joint visual and audio learning for video highlight detection.
\newblock In \emph{Proceedings of the IEEE/CVF International Conference on Computer Vision}, 8127--8137.

\bibitem[{Badamdorj et~al.(2022)Badamdorj, Rochan, Wang, and Cheng}]{badamdorj2022contrastive}
Badamdorj, T.; Rochan, M.; Wang, Y.; and Cheng, L. 2022.
\newblock Contrastive learning for unsupervised video highlight detection.
\newblock In \emph{Proceedings of the IEEE/CVF Conference on Computer Vision and Pattern Recognition}, 14042--14052.

\bibitem[{Escorcia et~al.(2019)Escorcia, Soldan, Sivic, Ghanem, and Russell}]{escorcia2019temporal}
Escorcia, V.; Soldan, M.; Sivic, J.; Ghanem, B.; and Russell, B. 2019.
\newblock Temporal localization of moments in video collections with natural language.

\bibitem[{Feichtenhofer et~al.(2019)Feichtenhofer, Fan, Malik, and He}]{feichtenhofer2019slowfast}
Feichtenhofer, C.; Fan, H.; Malik, J.; and He, K. 2019.
\newblock Slowfast networks for video recognition.
\newblock In \emph{Proceedings of the IEEE/CVF international conference on computer vision}, 6202--6211.

\bibitem[{Gao et~al.(2017)Gao, Sun, Yang, and Nevatia}]{gao2017tall}
Gao, J.; Sun, C.; Yang, Z.; and Nevatia, R. 2017.
\newblock Tall: Temporal activity localization via language query.
\newblock In \emph{Proceedings of the IEEE international conference on computer vision}, 5267--5275.

\bibitem[{Gao et~al.(2021)Gao, Sun, Xu, Zhou, and Ghanem}]{gao2021relation}
Gao, J.; Sun, X.; Xu, M.; Zhou, X.; and Ghanem, B. 2021.
\newblock Relation-aware Video Reading Comprehension for Temporal Language Grounding.
\newblock In \emph{Proceedings of the 2021 Conference on Empirical Methods in Natural Language Processing}, 3978--3988.

\bibitem[{Gao and Xu(2021)}]{gao2021fast}
Gao, J.; and Xu, C. 2021.
\newblock Fast video moment retrieval.
\newblock In \emph{Proceedings of the IEEE/CVF International Conference on Computer Vision}, 1523--1532.

\bibitem[{Hendricks and Nematzadeh(2021)}]{hendricks2021probing}
Hendricks, L.~A.; and Nematzadeh, A. 2021.
\newblock Probing image-language transformers for verb understanding.
\newblock \emph{arXiv preprint arXiv:2106.09141}.

\bibitem[{Hong et~al.(2020)Hong, Huang, Li, and Zheng}]{hong2020mini}
Hong, F.-T.; Huang, X.; Li, W.-H.; and Zheng, W.-S. 2020.
\newblock Mini-net: Multiple instance ranking network for video highlight detection.
\newblock In \emph{Computer Vision--ECCV 2020: 16th European Conference, Glasgow, UK, August 23--28, 2020, Proceedings, Part XIII 16}, 345--360. Springer.

\bibitem[{Houlsby et~al.(2019)Houlsby, Giurgiu, Jastrzebski, Morrone, De~Laroussilhe, Gesmundo, Attariyan, and Gelly}]{houlsby2019parameter}
Houlsby, N.; Giurgiu, A.; Jastrzebski, S.; Morrone, B.; De~Laroussilhe, Q.; Gesmundo, A.; Attariyan, M.; and Gelly, S. 2019.
\newblock Parameter-efficient transfer learning for NLP.
\newblock In \emph{International conference on machine learning}, 2790--2799. PMLR.

\bibitem[{Jia et~al.(2021)Jia, Yang, Xia, Chen, Parekh, Pham, Le, Sung, Li, and Duerig}]{jia2021scaling}
Jia, C.; Yang, Y.; Xia, Y.; Chen, Y.-T.; Parekh, Z.; Pham, H.; Le, Q.; Sung, Y.-H.; Li, Z.; and Duerig, T. 2021.
\newblock Scaling up visual and vision-language representation learning with noisy text supervision.
\newblock In \emph{International Conference on Machine Learning}, 4904--4916. PMLR.

\bibitem[{Jia et~al.(2022)Jia, Tang, Chen, Cardie, Belongie, Hariharan, and Lim}]{jia2022visual}
Jia, M.; Tang, L.; Chen, B.-C.; Cardie, C.; Belongie, S.; Hariharan, B.; and Lim, S.-N. 2022.
\newblock Visual prompt tuning.
\newblock \emph{arXiv preprint arXiv:2203.12119}.

\bibitem[{Khattak et~al.(2022)Khattak, Rasheed, Maaz, Khan, and Khan}]{khattak2022maple}
Khattak, M.~U.; Rasheed, H.; Maaz, M.; Khan, S.; and Khan, F.~S. 2022.
\newblock Maple: Multi-modal prompt learning.
\newblock \emph{arXiv preprint arXiv:2210.03117}.

\bibitem[{Lei, Berg, and Bansal(2021)}]{lei2021detecting}
Lei, J.; Berg, T.~L.; and Bansal, M. 2021.
\newblock Detecting moments and highlights in videos via natural language queries.
\newblock \emph{Advances in Neural Information Processing Systems}, 34: 11846--11858.

\bibitem[{Lei et~al.(2020)Lei, Yu, Berg, and Bansal}]{lei2020tvr}
Lei, J.; Yu, L.; Berg, T.~L.; and Bansal, M. 2020.
\newblock Tvr: A large-scale dataset for video-subtitle moment retrieval.
\newblock In \emph{Computer Vision--ECCV 2020: 16th European Conference, Glasgow, UK, August 23--28, 2020, Proceedings, Part XXI 16}, 447--463. Springer.

\bibitem[{Lester, Al-Rfou, and Constant(2021)}]{lester2021power}
Lester, B.; Al-Rfou, R.; and Constant, N. 2021.
\newblock The power of scale for parameter-efficient prompt tuning.
\newblock \emph{arXiv preprint arXiv:2104.08691}.

\bibitem[{Li, Guo, and Wang(2021)}]{li2021proposal}
Li, K.; Guo, D.; and Wang, M. 2021.
\newblock Proposal-free video grounding with contextual pyramid network.
\newblock In \emph{Proceedings of the AAAI Conference on Artificial Intelligence}, volume~35, 1902--1910.

\bibitem[{Li and Liang(2021)}]{li2021prefix}
Li, X.~L.; and Liang, P. 2021.
\newblock Prefix-tuning: Optimizing continuous prompts for generation.
\newblock \emph{arXiv preprint arXiv:2101.00190}.

\bibitem[{Lin et~al.(2023)Lin, Zhang, Chen, Pramanick, Gao, Wang, Yan, and Shou}]{lin2023univtg}
Lin, K.~Q.; Zhang, P.; Chen, J.; Pramanick, S.; Gao, D.; Wang, A.~J.; Yan, R.; and Shou, M.~Z. 2023.
\newblock Univtg: Towards unified video-language temporal grounding.
\newblock In \emph{Proceedings of the IEEE/CVF International Conference on Computer Vision}, 2794--2804.

\bibitem[{Liu et~al.(2021)Liu, Ji, Fu, Du, Yang, and Tang}]{liu2021p}
Liu, X.; Ji, K.; Fu, Y.; Du, Z.; Yang, Z.; and Tang, J. 2021.
\newblock P-tuning v2: Prompt tuning can be comparable to fine-tuning universally across scales and tasks.
\newblock \emph{arXiv preprint arXiv:2110.07602}.

\bibitem[{Liu et~al.(2022)Liu, Li, Wu, Chen, Shan, and Qie}]{liu2022umt}
Liu, Y.; Li, S.; Wu, Y.; Chen, C.-W.; Shan, Y.; and Qie, X. 2022.
\newblock Umt: Unified multi-modal transformers for joint video moment retrieval and highlight detection.
\newblock In \emph{Proceedings of the IEEE/CVF Conference on Computer Vision and Pattern Recognition}, 3042--3051.

\bibitem[{Lu et~al.(2022)Lu, Liu, Zhang, Liu, and Tian}]{lu2022prompt}
Lu, Y.; Liu, J.; Zhang, Y.; Liu, Y.; and Tian, X. 2022.
\newblock Prompt distribution learning.
\newblock In \emph{Proceedings of the IEEE/CVF Conference on Computer Vision and Pattern Recognition}, 5206--5215.

\bibitem[{Moon et~al.(2023)Moon, Hyun, Park, Park, and Heo}]{moon2023query}
Moon, W.; Hyun, S.; Park, S.; Park, D.; and Heo, J.-P. 2023.
\newblock Query-dependent video representation for moment retrieval and highlight detection.
\newblock In \emph{Proceedings of the IEEE/CVF Conference on Computer Vision and Pattern Recognition}, 23023--23033.

\bibitem[{Mun, Cho, and Han(2020)}]{mun2020local}
Mun, J.; Cho, M.; and Han, B. 2020.
\newblock Local-global video-text interactions for temporal grounding.
\newblock In \emph{Proceedings of the IEEE/CVF Conference on Computer Vision and Pattern Recognition}, 10810--10819.

\bibitem[{Radford et~al.(2021)Radford, Kim, Hallacy, Ramesh, Goh, Agarwal, Sastry, Askell, Mishkin, Clark et~al.}]{radford2021learning}
Radford, A.; Kim, J.~W.; Hallacy, C.; Ramesh, A.; Goh, G.; Agarwal, S.; Sastry, G.; Askell, A.; Mishkin, P.; Clark, J.; et~al. 2021.
\newblock Learning transferable visual models from natural language supervision.
\newblock In \emph{International conference on machine learning}, 8748--8763. PMLR.

\bibitem[{Regneri et~al.(2013)Regneri, Rohrbach, Wetzel, Thater, Schiele, and Pinkal}]{regneri2013grounding}
Regneri, M.; Rohrbach, M.; Wetzel, D.; Thater, S.; Schiele, B.; and Pinkal, M. 2013.
\newblock Grounding action descriptions in videos.
\newblock \emph{Transactions of the Association for Computational Linguistics}, 1: 25--36.

\bibitem[{Song et~al.(2015)Song, Vallmitjana, Stent, and Jaimes}]{song2015tvsum}
Song, Y.; Vallmitjana, J.; Stent, A.; and Jaimes, A. 2015.
\newblock Tvsum: Summarizing web videos using titles.
\newblock In \emph{Proceedings of the IEEE conference on computer vision and pattern recognition}, 5179--5187.

\bibitem[{Sun et~al.(2024)Sun, Zhou, Chen, and Xie}]{sun2024tr}
Sun, H.; Zhou, M.; Chen, W.; and Xie, W. 2024.
\newblock Tr-detr: Task-reciprocal transformer for joint moment retrieval and highlight detection.
\newblock In \emph{Proceedings of the AAAI Conference on Artificial Intelligence}, volume~38, 4998--5007.

\bibitem[{Sun, Farhadi, and Seitz(2014)}]{sun2014ranking}
Sun, M.; Farhadi, A.; and Seitz, S. 2014.
\newblock Ranking domain-specific highlights by analyzing edited videos.
\newblock In \emph{Computer Vision--ECCV 2014: 13th European Conference, Zurich, Switzerland, September 6-12, 2014, Proceedings, Part I 13}, 787--802. Springer.

\bibitem[{Wang et~al.(2023)Wang, Jiang, Cheng, Li, and Zhao}]{wang2023learning}
Wang, Y.; Jiang, X.; Cheng, D.; Li, D.; and Zhao, C. 2023.
\newblock Learning Hierarchical Prompt with Structured Linguistic Knowledge for Vision-Language Models.
\newblock \emph{arXiv preprint arXiv:2312.06323}.

\bibitem[{Wei et~al.(2022)Wei, Wang, Ge, Jiang, Li, and Duan}]{wei2022learning}
Wei, F.; Wang, B.; Ge, T.; Jiang, Y.; Li, W.; and Duan, L. 2022.
\newblock Learning pixel-level distinctions for video highlight detection.
\newblock In \emph{Proceedings of the IEEE/CVF Conference on Computer Vision and Pattern Recognition}, 3073--3082.

\bibitem[{Xiao et~al.(2021)Xiao, Chen, Zhang, Ji, Shao, Ye, and Xiao}]{xiao2021boundary}
Xiao, S.; Chen, L.; Zhang, S.; Ji, W.; Shao, J.; Ye, L.; and Xiao, J. 2021.
\newblock Boundary proposal network for two-stage natural language video localization.
\newblock In \emph{Proceedings of the AAAI Conference on Artificial Intelligence}, volume~35, 2986--2994.

\bibitem[{Xu et~al.(2019)Xu, He, Plummer, Sigal, Sclaroff, and Saenko}]{xu2019multilevel}
Xu, H.; He, K.; Plummer, B.~A.; Sigal, L.; Sclaroff, S.; and Saenko, K. 2019.
\newblock Multilevel language and vision integration for text-to-clip retrieval.
\newblock In \emph{Proceedings of the AAAI Conference on Artificial Intelligence}, volume~33, 9062--9069.

\bibitem[{Xu et~al.(2021)Xu, Wang, Ni, Zhu, Sun, and Wang}]{xu2021cross}
Xu, M.; Wang, H.; Ni, B.; Zhu, R.; Sun, Z.; and Wang, C. 2021.
\newblock Cross-category video highlight detection via set-based learning.
\newblock In \emph{Proceedings of the IEEE/CVF International Conference on Computer Vision}, 7970--7979.

\bibitem[{Xu et~al.(2022)Xu, Chen, Wei, Deng, and Xue}]{xu2022hisa}
Xu, Z.; Chen, D.; Wei, K.; Deng, C.; and Xue, H. 2022.
\newblock HiSA: Hierarchically semantic associating for video temporal grounding.
\newblock \emph{IEEE Transactions on Image Processing}, 31: 5178--5188.

\bibitem[{Yan et~al.(2023)Yan, Xiong, Nagrani, Arnab, Wang, Ge, Ross, and Schmid}]{yan2023unloc}
Yan, S.; Xiong, X.; Nagrani, A.; Arnab, A.; Wang, Z.; Ge, W.; Ross, D.; and Schmid, C. 2023.
\newblock Unloc: A unified framework for video localization tasks.
\newblock In \emph{Proceedings of the IEEE/CVF International Conference on Computer Vision}, 13623--13633.

\bibitem[{Zhang et~al.(2019)Zhang, Dai, Wang, Wang, and Davis}]{zhang2019man}
Zhang, D.; Dai, X.; Wang, X.; Wang, Y.-F.; and Davis, L.~S. 2019.
\newblock Man: Moment alignment network for natural language moment retrieval via iterative graph adjustment.
\newblock In \emph{Proceedings of the IEEE/CVF Conference on Computer Vision and Pattern Recognition}, 1247--1257.

\bibitem[{Zhang et~al.(2020{\natexlab{a}})Zhang, Sun, Jing, and Zhou}]{zhang2020span}
Zhang, H.; Sun, A.; Jing, W.; and Zhou, J.~T. 2020{\natexlab{a}}.
\newblock Span-based Localizing Network for Natural Language Video Localization.
\newblock In \emph{Proceedings of the 58th Annual Meeting of the Association for Computational Linguistics}, 6543--6554.

\bibitem[{Zhang et~al.(2020{\natexlab{b}})Zhang, Peng, Fu, and Luo}]{zhang2020learning}
Zhang, S.; Peng, H.; Fu, J.; and Luo, J. 2020{\natexlab{b}}.
\newblock Learning 2d temporal adjacent networks for moment localization with natural language.
\newblock In \emph{Proceedings of the AAAI Conference on Artificial Intelligence}, volume~34, 12870--12877.

\bibitem[{Zhao et~al.(2024)Zhao, Wang, Jiang, Shen, Song, Li, and Miao}]{zhao2024learning}
Zhao, C.; Wang, Y.; Jiang, X.; Shen, Y.; Song, K.; Li, D.; and Miao, D. 2024.
\newblock Learning domain invariant prompt for vision-language models.
\newblock \emph{IEEE Transactions on Image Processing}.

\bibitem[{Zhou et~al.(2022{\natexlab{a}})Zhou, Yang, Loy, and Liu}]{zhou2022conditional}
Zhou, K.; Yang, J.; Loy, C.~C.; and Liu, Z. 2022{\natexlab{a}}.
\newblock Conditional prompt learning for vision-language models.
\newblock In \emph{Proceedings of the IEEE/CVF Conference on Computer Vision and Pattern Recognition}, 16816--16825.

\bibitem[{Zhou et~al.(2022{\natexlab{b}})Zhou, Yang, Loy, and Liu}]{zhou2022learning}
Zhou, K.; Yang, J.; Loy, C.~C.; and Liu, Z. 2022{\natexlab{b}}.
\newblock Learning to prompt for vision-language models.
\newblock \emph{International Journal of Computer Vision}, 130(9): 2337--2348.

\bibitem[{Zhu et~al.(2023)Zhu, Niu, Han, Wu, and Zhang}]{zhu2023prompt}
Zhu, B.; Niu, Y.; Han, Y.; Wu, Y.; and Zhang, H. 2023.
\newblock Prompt-aligned gradient for prompt tuning.
\newblock In \emph{Proceedings of the IEEE/CVF International Conference on Computer Vision}, 15659--15669.

\end{thebibliography}

\end{document}